\documentclass[conference]{IEEEtran}
\usepackage{graphicx}
\usepackage{booktabs}
\usepackage{amsmath}
\usepackage{url}
\usepackage[hidelinks]{hyperref}

\begin{document}

\title{Auditing Demonstration Curation Metrics: Action-Only Scorers Fail on the
Structural Defects That Degrade Imitation Policies}

\author{\IEEEauthorblockN{Aarav Bedi}
\IEEEauthorblockA{Department of Mechanical Engineering\\
University of California, Berkeley\\
Berkeley, California, USA\\
aaravbedi@berkeley.edu}}

\maketitle

\begin{abstract}
Imitation-learning policies inherit the quality of the demonstrations they are
trained on, and a growing set of curation metrics promise to score and filter
low-quality demonstrations automatically. These metrics are each validated on
different data with different protocols, so it is unclear which of them actually
identify the demonstrations that harm a policy. We build a controlled testbed in
which demonstration defects are injected with known type, and audit seven
curation metrics along two axes: how well each separates defective from clean
demonstrations, and whether training a behavior-cloning policy on each metric's
curated subset improves task success. We study two defect regimes. Subtle
perturbations (correlated action noise, tremor, truncation) are detectable by
multivariate outlier scoring and, once removed, recover the full downstream gap.
Structural errors, where the demonstration executes a wrong action at a key
moment, are invisible to every action-only metric we test, and two of them are
inverted: they score defective demonstrations as higher quality and, used for
curation, tend to leave the policy at or below the uncurated baseline rather than
above it. Only metrics that examine the state trajectory detect structural errors,
and even the best of them recovers just a third of the downstream gap. High
detection accuracy does not guarantee downstream improvement. We release the
testbed and all curation implementations.
\end{abstract}


\section{Introduction}
A policy trained by imitation can only be as good as the demonstrations it
copies. Large demonstration sets are rarely uniform in quality: operators differ
in skill, scripted controllers occasionally fail, and autonomous rollouts get
labeled successful by checks that are themselves imperfect. Whatever the source,
some of the data is bad, and a behavior-cloning policy has no way to tell the
good demonstrations from the bad. It imitates all of them.

Curation is the obvious fix. Score each demonstration with some quality metric,
drop the low-scoring ones, and train on what remains. The literature offers no
shortage of metrics for this, from trajectory smoothness to outlier detection in
a feature space. What it does not offer is a basis for choosing between them:
each is reported on its own dataset under its own protocol. There is also a
subtler problem. A curation metric is rewarded for telling you which
demonstrations look unusual, but what you care about is which demonstrations hurt
the policy. Those are different questions, and nothing guarantees that a metric
good at the first is any good at the second.

We separate the two questions in a controlled testbed. We simulate a
manipulation task, inject defects into demonstrations with a known type and
location, and withhold those labels from every metric. Each of seven metrics is
then scored two ways. The first is detection: how cleanly its quality ranking
separates defective demonstrations from clean ones. The second is downstream:
whether a behavior-cloning policy trained on the subset it keeps actually
succeeds more often when run in the simulator. The second number is the one a
practitioner cares about, and by construction it does not depend on any curation
score.

We study two regimes of demonstration defect. In the first, the defect is a
subtle perturbation of an otherwise correct demonstration: correlated action
noise, a high-frequency tremor, or an early truncation. In the second, the
defect is structural: the demonstration executes a wrong action at a decisive
moment, releasing the grasped object partway through the task. Our findings are:

\begin{itemize}
\item The two regimes have opposite detection--degradation profiles. Subtle
perturbations are detectable and, once removed, the downstream gap closes
completely. Structural errors cause a far larger downstream gap and almost
nothing recovers it.
\item Action-only metrics can be not merely uninformative but inverted. On the
structural defect, two metrics score defective demonstrations as higher quality
than clean ones; used for curation they yield policies worse than training on
the uncurated contaminated set.
\item Detecting a defect does not guarantee recovering from it. One metric
detects the structural defect well yet its curated policy barely beats the
no-curation baseline, while another with similar detection accuracy recovers a
third of the gap.
\end{itemize}

Our contribution is the audit and the testbed, not a new curation method. We are
after a simple thing: with the trained policy as the judge, which of these
metrics actually help, which do nothing, and which make matters worse.

\section{Related Work}
\textbf{Imitation learning from demonstrations.} Behavior cloning trains a
policy by supervised regression from observations to demonstrated
actions~\cite{alvinn}, and remains the backbone of modern manipulation learning
despite its known sensitivity to distribution shift~\cite{dagger}. Recent
bimanual systems such as ACT collect and imitate large demonstration sets on
low-cost hardware~\cite{act}, and large cross-embodiment corpora and standard
data formats have scaled this further~\cite{oxe,bridge,droid,lerobot}. All of
these assume the demonstrations are worth imitating.

\textbf{Demonstration quality and curation.} The observation that demonstration
quality, not just quantity, drives imitation performance is central to studies
of offline human data~\cite{robomimic}, which document large differences between
operators. This motivates automatic curation: scoring demonstrations and
filtering low-quality ones. Proposed signals range from trajectory smoothness,
for which spectral arc length is a standard movement-quality
measure~\cite{sparc}, to generic outlier detection such as isolation
forests~\cite{iforest}. What has been missing is a controlled, head-to-head
comparison that asks whether these signals identify the demonstrations that
actually degrade a policy, rather than merely the ones that look unusual.

\section{Testbed and Protocol}
\textbf{Environment.} We use a lightweight pick-and-place simulator written in
NumPy, modeled on the single-arm version of the ALOHA setup. The action is
seven-dimensional (end-effector translation, rotation, and gripper). The
behavior-cloning observation is eleven-dimensional: end-effector position and
orientation, gripper state, a noisy estimate of the object position (Gaussian
noise with standard deviation $0.03$\,m applied at every step, standing in for
imperfect perception), and a normalized time index. The task is to grasp the
object and carry it to a fixed goal region; success is the object resting in the
goal at episode end. A phase-based scripted controller with access to privileged
state solves the task essentially every time and supplies the clean
demonstrations; the learned policy never sees that privileged state.

\textbf{Two labels, kept apart.} A clean demonstration is a successful scripted
episode. A defective demonstration is produced by applying one of the injectors
below to a clean episode; its defect type is recorded but is never made
available to any curation metric. Curation metrics see only a view exposing the
states and actions of a demonstration, enforced at the type level so a metric
cannot accidentally read the label.

\textbf{Defect regimes.} We study two regimes, each at a $40\%$ contamination
rate.
\emph{Subtle perturbations:} (i) \textbf{action noise}, temporally correlated
AR(1) noise added to the actions, which looks like shaky teleoperation rather
than random garbage; (ii) \textbf{tremor}, a high-frequency sinusoid added to
the actions; (iii) \textbf{truncation}, cutting the episode to roughly half its
length; (iv) \textbf{detour}, splicing a reversed copy of a mid-episode segment
back in so the trajectory loops before completing.
\emph{Structural error:} \textbf{early release}, the gripper is commanded open
during the carry phase, so the demonstration drops the object partway to the
goal. This is not noise; it is a systematic wrong action in a specific part of
the state space, and it does not average out across demonstrations.

\textbf{Curation metrics.} We audit seven metrics, each producing a per-demo
quality score from states and actions only.
\emph{Action-only:} \textbf{smoothness} (spectral arc length of the movement
speed profile), \textbf{entropy} (standard deviation of the action sequence),
\textbf{length} (trajectory length), \textbf{isolation forest} and
\textbf{ensemble}, both operating on a vector of action-derived summary
features.
\emph{State-trajectory-aware:} \textbf{kNN}, scoring a demonstration by its
distance to its nearest neighbors in a trajectory-level feature space that
includes state-trajectory summaries, so demonstrations that leave the local
data manifold score low; and \textbf{trajectory alignment}, scoring how well a
demonstration's state trajectory agrees with the dataset's aggregate behavior.
An \textbf{oracle} that filters using the hidden labels upper-bounds what
perfect curation could achieve.

\textbf{Evaluation.} On the detection axis we report the area under the ROC
curve (AUROC) for each metric's quality score against the hidden defective/clean
label. On the downstream axis we curate the contaminated set by keeping the
top-scoring fraction under each metric, train a three-layer behavior-cloning MLP
on the kept subset, and measure task success over $50$ fresh simulator rollouts.
To remove a confound between data quantity and data quality, every downstream
condition trains on exactly $N=150$ demonstrations sampled from its kept pool, so
differences reflect quality rather than volume. We report mean and standard
deviation over three seeds ($42$, $0$, $7$).

\section{Results}
\begin{figure}[t]
\centering
\includegraphics[width=\columnwidth]{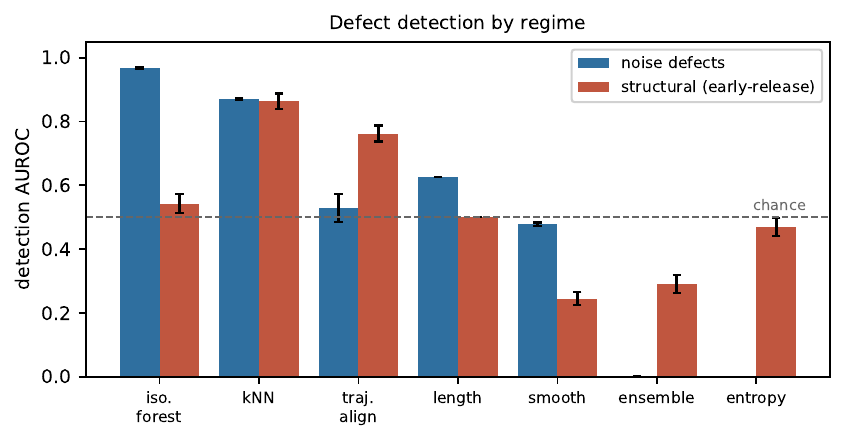}
\caption{Defect-detection AUROC by metric and regime. No metric is reliable in
both regimes; entropy and ensemble are inverted (below chance) on one or both.}
\label{fig:auroc}
\end{figure}

\textbf{Detection flips between regimes.} Fig.~\ref{fig:auroc} and
Tables~\ref{tab:auroc_noise} and~\ref{tab:auroc_struct} show that no metric is
good in both regimes. On subtle perturbations the multivariate isolation forest
is strongest (AUROC $0.97$) and the state-trajectory kNN is close behind; on the
structural defect the isolation forest collapses to chance ($0.54$) while kNN
and trajectory alignment, which read the state trajectory, are the only metrics
that detect it. The smoothness and entropy metrics fall below chance on the
structural defect, and entropy is essentially inverted on the subtle regime: the
correlated noise and tremor both inflate action variance, so entropy scores the
defective demonstrations as more exploratory, hence higher quality, than the
clean ones. The ensemble inherits this inversion because the entropy feature
drags its composite score the wrong way.

\begin{table}[t]
\caption{Detection AUROC, subtle-perturbation regime (3 seeds)}
\label{tab:auroc_noise}
\centering
\begin{tabular}{lc}
\toprule
Metric & AUROC \\
\midrule
isolation forest      & $0.968 \pm 0.004$ \\
kNN                   & $0.871 \pm 0.003$ \\
length                & $0.625 \pm 0.000$ \\
trajectory alignment  & $0.528 \pm 0.044$ \\
smoothness            & $0.478 \pm 0.005$ \\
ensemble              & $0.002 \pm 0.001$ \\
entropy               & $0.000 \pm 0.000$ \\
\bottomrule
\end{tabular}
\end{table}

\begin{table}[t]
\caption{Detection AUROC, structural-defect regime (3 seeds)}
\label{tab:auroc_struct}
\centering
\begin{tabular}{lc}
\toprule
Metric & AUROC \\
\midrule
kNN                   & $0.863 \pm 0.025$ \\
trajectory alignment  & $0.762 \pm 0.025$ \\
isolation forest      & $0.543 \pm 0.031$ \\
length                & $0.500 \pm 0.000$ \\
entropy               & $0.469 \pm 0.028$ \\
ensemble              & $0.291 \pm 0.028$ \\
smoothness            & $0.244 \pm 0.020$ \\
\bottomrule
\end{tabular}
\end{table}

\textbf{Subtle defects are recoverable.} In the subtle-perturbation regime
(Table~\ref{tab:downstream_noise}) the contaminated set trains a policy at
$55.3\%$ against an oracle ceiling of $72.0\%$, and the isolation forest recovers
most of that gap ($71.3\%$). The downstream spread in this regime is wide across
seeds, though: with three seeds the per-method standard deviations run as high as
$\pm 20$ points, so we read these subtle-regime downstream magnitudes as
suggestive and lean on the detection axis, where the separation is sharp and
stable, for the firmer statements. The clearest of those is entropy's inversion
(AUROC $0.000 \pm 0.000$ across seeds): it ranks the noisier demonstrations as
higher quality, the opposite of what curation needs, because the correlated noise
and tremor both inflate action variance.

\begin{table}[t]
\caption{Downstream success, subtle-perturbation regime ($N{=}150$, 50 rollouts,
3 seeds)}
\label{tab:downstream_noise}
\centering
\begin{tabular}{lc}
\toprule
Training set & Success (\%) \\
\midrule
oracle clean only      & $72.0 \pm 4.9$ \\
isolation forest       & $71.3 \pm 10.9$ \\
smoothness             & $70.0 \pm 9.1$ \\
length                 & $56.7 \pm 15.5$ \\
ensemble               & $56.0 \pm 16.1$ \\
full contaminated      & $55.3 \pm 0.9$ \\
entropy                & $41.3 \pm 20.2$ \\
\bottomrule
\end{tabular}
\end{table}

\begin{figure}[t]
\centering
\includegraphics[width=\columnwidth]{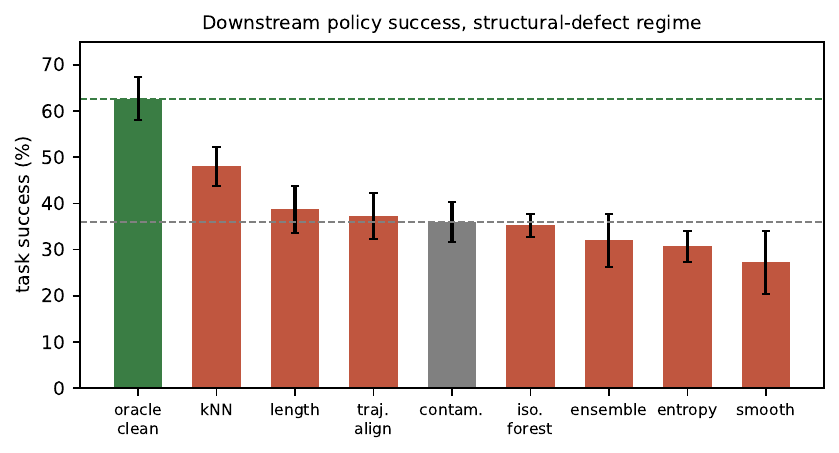}
\caption{Downstream policy success in the structural-defect regime. Dashed lines
mark the oracle (green) and no-curation (gray) references. Only kNN improves
meaningfully on the contaminated baseline; action-only metrics match it or, for
smoothness, fall below it.}
\label{fig:downstream}
\end{figure}

\textbf{Structural defects resist curation.} The structural regime is where
curation matters most and helps least (Table~\ref{tab:downstream_struct},
Fig.~\ref{fig:downstream}). The contaminated policy succeeds only $36.0\%$ of
the time against a $62.7\%$ oracle, a $27$-point gap. The one clear improvement
is kNN, which reaches $48.0 \pm 4.3\%$ and recovers about a third of the gap,
roughly two standard deviations clear of the baseline across seeds. No other
metric improves on the contaminated set. The inverted action-only metrics
(smoothness, ensemble, entropy) all sit below the baseline in point estimate,
smoothness lowest at $27.3\%$, but with three seeds these below-baseline gaps
fall within seed variance, so we report them as a consistent trend rather than a
sharp effect. The robust statement is that no action-only metric helps here,
while the one state-trajectory metric that helps does so only partially.

\begin{table}[t]
\caption{Downstream success, structural-defect regime ($N{=}150$, 50 rollouts,
3 seeds)}
\label{tab:downstream_struct}
\centering
\begin{tabular}{lc}
\toprule
Training set & Success (\%) \\
\midrule
oracle clean only      & $62.7 \pm 4.7$ \\
kNN                    & $48.0 \pm 4.3$ \\
length                 & $38.7 \pm 5.0$ \\
trajectory alignment   & $37.3 \pm 5.0$ \\
full contaminated      & $36.0 \pm 4.3$ \\
isolation forest       & $35.3 \pm 2.5$ \\
ensemble               & $32.0 \pm 5.7$ \\
entropy                & $30.7 \pm 3.4$ \\
smoothness             & $27.3 \pm 6.8$ \\
\bottomrule
\end{tabular}
\end{table}

\begin{figure}[t]
\centering
\includegraphics[width=0.86\columnwidth]{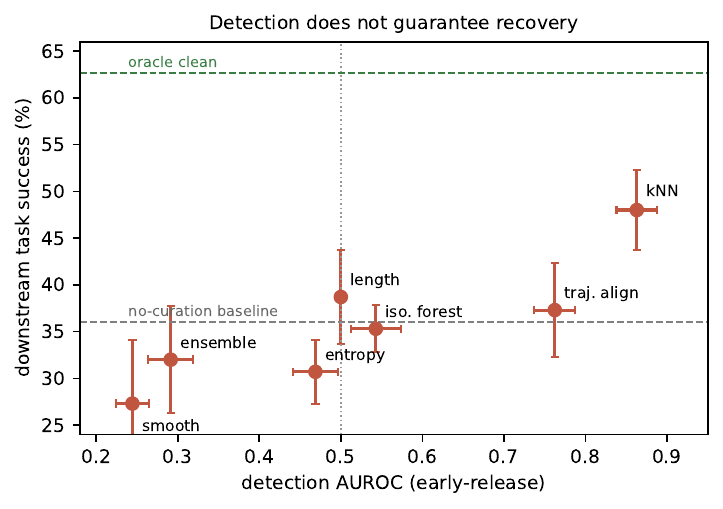}
\caption{Detection AUROC versus downstream success in the structural regime.
Comparable detection (kNN, trajectory alignment) yields very different
downstream recovery.}
\label{fig:scatter}
\end{figure}

\textbf{Detection does not guarantee recovery.} Fig.~\ref{fig:scatter} plots
structural-defect detection against downstream success. The two state-aware
metrics detect the defect comparably well (AUROC $0.86$ and $0.76$), yet kNN
recovers a third of the downstream gap while trajectory alignment barely clears
the no-curation baseline. Detection accuracy is necessary but not sufficient:
which demonstrations a metric removes, and which clean ones it discards as
collateral, matters as much as its ranking quality. The plausible reason is that
trajectory alignment scores against the dataset's aggregate behavior, which the
$60\%$ clean majority pulls toward clean trajectories, diluting the signal,
whereas kNN's local-neighborhood comparison is less diluted.

\section{Discussion}
Two things come out of this audit. The first is practical: match the curation
metric to the kind of defect you expect. Multivariate outlier scoring handles
subtle perturbations well, but only metrics that look at the state trajectory
have a chance on structural errors, and the action-only smoothness and entropy
scores can do real harm, throwing out good demonstrations and leaving a policy
worse than if it had not been curated at all. The second is a warning about how
curation methods are evaluated. A metric's detection score says little about its
value: two metrics with nearly identical AUROC on the structural defect produced
policies eleven points apart, and the metric that dominates the subtle regime,
isolation forest, is no better than chance on the structural one. Reporting a
curation method by its detection accuracy alone reports the wrong number. The
trained policy is the only judge that counts.

Why do the two regimes behave so differently? Behavior cloning averages over its
demonstrations, and the two defect families sit on opposite sides of that
average. The subtle perturbations are roughly zero-mean wobble layered on top of
otherwise correct behavior, so they wash out as demonstrations accumulate, and
even rough curation is enough. An early release is not zero-mean. It plants a
specific wrong action in a specific part of the state space, and averaging never
removes it. The defects worth catching are the ones whose signature lives in
where the arm went, not in the statistics of how it moved, and that is exactly
the information the action-only metrics discard.

\section{Limitations}
The scope here is narrow by design, and the results should be read with that in
mind. The simulator is a lightweight NumPy proxy rather than a contact-rich
physics engine, and the task is a single single-arm pick-and-place. The
behavior-cloning observation includes a noisy object-position estimate that
stands in for perception instead of raw pixels. The defects are injected with a
known type rather than harvested from real operators, and we test a single policy
class. These are the choices that bought a clean, label-controlled comparison,
and they cost realism. Two further caveats bear on how firmly the numbers should
be read. First, we run only three seeds, and several downstream estimates carry
large seed-to-seed variance, most of all in the subtle regime where per-method
standard deviations reach twenty points; we therefore treat the downstream
magnitudes as preliminary and the detection results, whose variance is small, as
the more stable signal. Second, because the defects are injected, the two
families were chosen on purpose to straddle the action/state divide, one visible
in the action statistics and one only in the state trajectory, so the action-only
metrics' blindness to the structural defect is in part by design. The substantive
results are the sizes of the downstream gaps and the finding that detecting a
defect did not let any metric recover from it, not the bare existence of a
state-only failure. The obvious next steps are many more seeds with confidence
intervals, a contact-rich benchmark such as robosuite or LIBERO, real
multi-operator quality labels of the kind found in offline human demonstration
datasets, more tasks and defect types, and policies beyond behavior cloning. We
expect the qualitative story to hold, structural errors slipping past action-only
curation and detection failing to imply recovery, but the numbers themselves will
move.

\section{Conclusion}
We audited seven demonstration-curation metrics in a testbed that pulls apart
two things usually conflated: whether a metric flags a defect, and whether
removing what it flags actually helps the policy. Subtle perturbations turn out
to be both detectable and recoverable. Structural errors are neither. They slip
past every action-only metric, and those metrics, used for curation, do not help
and in point estimate trend below the no-curation baseline. The scorer that wins
one regime sits at one regime sits at
chance in the other, and two scorers with matching detection scores can land far
apart downstream. If there is one thing to carry away, it is to judge a curation
method by the policy it produces rather than the defects it flags, and to reach
for state-trajectory-aware metrics when the errors that matter are structural.

\section*{Data and Code Availability}
The simulator, defect injectors, all seven curation metrics, and the evaluation
pipeline that produces every table and figure in this paper are available at
\url{https://github.com/aaravbedi/scorer-fail-on-structural-defects}.

\section*{Acknowledgment}
The author used Anthropic's Claude to assist with drafting and editing this
manuscript. The study was designed by the author, and all experiments were run
and all results verified by the author, who takes full responsibility for the
content.

\end{document}